%% file: acl_latex.tex
\pdfoutput=1

\documentclass[11pt]{article}

\usepackage[preprint]{acl}

\usepackage{times}
\usepackage{latexsym}

\usepackage[T1]{fontenc}

\usepackage[utf8]{inputenc}

\usepackage{microtype}

\usepackage{inconsolata}

\usepackage{graphicx}

\input{math_commands.tex}

\usepackage{url}

\usepackage{microtype}
\usepackage{graphicx}

\usepackage{amsmath}
\usepackage{amssymb}
\usepackage{mathtools}
\usepackage{amsthm}
\usepackage[capitalize,noabbrev]{cleveref}

\theoremstyle{plain}

\theoremstyle{definition}

\theoremstyle{remark}

\usepackage[textsize=tiny]{todonotes}
\PassOptionsToPackage{hyphens}{url}
\usepackage{url}

\usepackage{hyperref}
\usepackage[hyphens]{url}
\usepackage{xurl}  
\usepackage{url}
\usepackage{graphicx}%
\usepackage{multirow}%
\usepackage{mathrsfs}%
\usepackage{xcolor}%
\usepackage{textcomp}%
\usepackage{manyfoot}%
\usepackage{booktabs}%
\usepackage{algorithm}%
\usepackage{listings}%

\usepackage{physics}
\usepackage{fontawesome}
\usepackage{float}

\usepackage[utf8]{inputenc} 
\usepackage[T1]{fontenc}    
\usepackage{amsfonts}       
\usepackage{nicefrac}       
\usepackage{microtype}      

\usepackage[utf8]{inputenc} 
\usepackage[T1]{fontenc}    
\usepackage{url}            
\usepackage{amsfonts}       
\usepackage{nicefrac}       
\usepackage{bm}
\usepackage{microtype}      
\usepackage{xcolor}         
\usepackage{pifont}
\usepackage{graphicx}
\usepackage{subcaption}

\usepackage[export]{adjustbox}
\usepackage[framemethod=tikz]{mdframed}

\usepackage{float}
\usepackage{multirow}
\usepackage{wrapfig}
\usepackage[T1]{fontenc}
\usepackage[utf8]{inputenc}
\usepackage{babel}
\usepackage[font=small,labelfont=bf]{caption}
\usepackage{xspace}
\usepackage{algorithm}
\usepackage{algorithmic}
\definecolor{darkgreen}{rgb}{0,0.7,0.5}

\usepackage{tcolorbox}
\tcbuselibrary{skins}
\usepackage{tcolorbox}
\usepackage{xcolor}
\definecolor{myblue}{rgb}{0.2,0.2,0.6}
\tcbset{
    dialogstyle/.style={
        enhanced,
        colback=white,
        colframe=gray!30, 
        fonttitle=\bfseries,
        sharp corners,
        title={#1},
        attach title to upper,
        width=0.6\textwidth,
        boxrule=0.5mm, 
        top=8mm, 
    }
}

\newcommand{\llm}{LLM\xspace}
\newcommand{\llms}{LLMs\xspace}

\usepackage{colortbl}

\definecolor{Gray}{gray}{0.9}
\definecolor{LightCyan}{rgb}{0.88,1,1}

\newcommand{\dataset}{\texttt{\textbf{FRAMES}}\xspace}
\newcommand{\datasetF}{\texttt{\textbf{F}}actuality, \texttt{\textbf{R}}etrieval, And reasoning \texttt{\textbf{ME}}asurement \texttt{\textbf{S}}et\xspace}
\newcommand{\gemini}{\texttt{Gemini-Pro-1.5-0514}\xspace}
\newcommand{\geminiF}{\texttt{Gemini-Flash-1.5-0514}\xspace}
\newcommand{\size}{824\xspace}

\newcommand{\gemmaL}{\texttt{Gemma2-27b}\xspace}

\usepackage[export]{adjustbox}
\usepackage[framemethod=tikz]{mdframed}
\newcommand\cmark {\textcolor{green}{\ding{52}}}
\newcommand\xmark {\textcolor{red}{\ding{55}}}

\usepackage{caption}
\title{Fact, Fetch, and Reason: A Unified Evaluation of \\ Retrieval-Augmented Generation}

\author{
 \textbf{Satyapriya Krishna\thanks{Work done at Google DeepMind.}\textsuperscript{1}},
 \textbf{Kalpesh Krishna\thanks{Equal contribution as internship hosts}\textsuperscript{2}},
 \textbf{Anhad Mohananey\textsuperscript{†}\textsuperscript{2}},
 \textbf{Steven Schwarcz\textsuperscript{2}},
\\
 \textbf{Adam Stambler\textsuperscript{2}},
 \textbf{Shyam Upadhyay\textsuperscript{2}},
 \textbf{Manaal Faruqui\textsuperscript{*}\textsuperscript{3}}
\\
 \textsuperscript{1}Harvard University,
 \textsuperscript{2}Google DeepMind,
 \textsuperscript{3}Meta
\\
 \small{
    \textbf{Correspondence:} \href{mailto:skrishna@g.harvard.edu}{skrishna@g.harvard.edu}
 }
}

\begin{document}
\maketitle
\pagestyle{empty}     
\begin{abstract}
Large Language Models (\llms) have shown significant improvements across cognitive tasks, with an emerging application in enhancing retrieval-augmented generation (RAG) capabilities. These systems require \llms to understand queries, retrieve relevant information, and synthesize accurate responses. Given their increasing real-world deployment, comprehensive evaluation is crucial. We propose \dataset (\datasetF), a high-quality dataset designed to test \llms' factual responses, retrieval capabilities, and reasoning in generating final answers. Unlike previous work evaluating these abilities in isolation, \dataset offers a unified framework for assessing \llm performance in end-to-end RAG scenarios. Our dataset comprises challenging multi-hop questions requiring integration of information from multiple sources. Baseline results show that even state-of-the-art \llms struggle, achieving 0.408 accuracy without retrieval. However, our proposed multi-step retrieval pipeline significantly improves accuracy to 0.66 (>50\% improvement). We aim to bridge evaluation gaps and assist in developing more robust RAG systems.
\end{abstract}

\input{010intro}

\input{030dataset}
\input{040experiments}

\input{020relwork}
\input{050conclusions}

\bibliography{custom}

\appendix
\input{appendix}


\end{document}

%% file: math_commands.tex
\usepackage{amsmath,amsfonts,bm}

\def\eqref#1{equation~\ref{#1}}

\def\1{\bm{1}}

\DeclareMathAlphabet{\mathsfit}{\encodingdefault}{\sfdefault}{m}{sl}
\SetMathAlphabet{\mathsfit}{bold}{\encodingdefault}{\sfdefault}{bx}{n}



%% file: 010intro.tex
\section{Introduction}
\label{sec:intro}
\vspace{-1mm}

Recent advancements in Large Language Models (\llms) have significantly enhanced their capabilities across various natural language processing tasks, especially in systems that demand both factual accuracy and sophisticated reasoning for complex queries \citep{zhao2023survey,DBLP:journals/corr/abs-2402-06625}. Retrieval-augmented generation (RAG) techniques \citep{lewis2020retrieval,fan2019eli5,guu2020retrieval} have become a powerful approach by leveraging the strengths of retrieval systems and the generative capabilities of \llms. These techniques are particularly effective for tasks requiring multi-hop reasoning, factual grounding, and synthesizing information from diverse knowledge domains \citep{gao2023retrieval}. However, despite this progress, the evaluation of RAG systems remains fragmented and insufficient, as existing benchmarks typically assess components like retrieval, factual correctness, and reasoning in isolation \citep{Yu2024EvaluationOR}. This piecemeal approach fails to capture the holistic performance of these systems in real-world applications \citep{yu2024evaluation}.

\begin{figure*}[t!]
    \centering
    \includegraphics[width=0.9\textwidth]{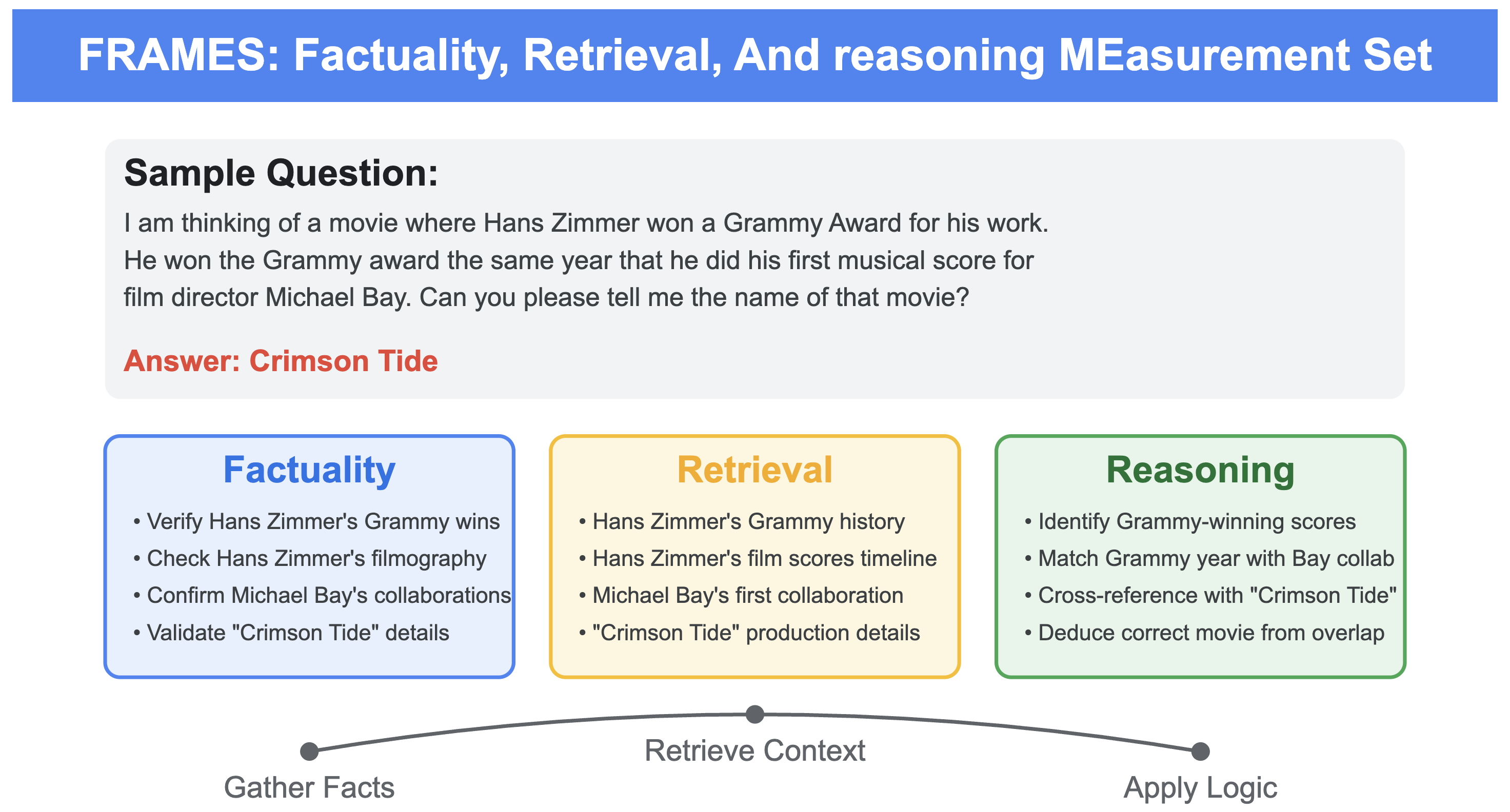}
    \caption{An example from the \dataset dataset, highlighting the core capabilities needed by a system (Factuality, Retrieval, Reasoning) to answer the question.}
    \label{fig:example}
\end{figure*}

To bridge this gap, we introduce a novel evaluation framework, \dataset \footnote{Dataset link : \url{https://huggingface.co/datasets/google/frames-benchmark}} (\datasetF), designed to rigorously test LLMs on all three core capabilities—fact retrieval, reasoning across multiple constraints, and accurate synthesis of information into coherent responses. Unlike existing datasets such as TruthfulQA \citep{lin2021truthfulqa}, HotpotQA \citep{yang2018hotpotqa}, or GSM8k \citep{cobbe2021gsm8k}, which focus on isolated aspects of LLM performance, \dataset provides an integrated evaluation that challenges models across these dimensions simultaneously. This approach offers a more accurate reflection of how these systems perform as end-to-end reasoning solutions, especially in scenarios requiring multi-document retrieval and complex reasoning. For example, our dataset includes questions like: \textit{"How many years earlier would Punxsutawney Phil have to be canonically alive to have made a Groundhog Day prediction in the same state as the US capitol?"} This demands temporal and numerical reasoning based on information from multiple retrieved articles. Figure \ref{fig:example} provides another such example.

Our work addresses a critical void in the current landscape by offering a challenging evaluation benchmark that not only tests the individual components of LLMs but also evaluates their performance in an end-to-end context. Through our dataset, we simulate realistic, multi-document queries to assess the ability of LLMs to retrieve relevant facts, reason accurately, and synthesize information into coherent responses. Additionally, we present empirical results on the performance of state-of-the-art models, highlighting both their strengths and the limitations in their reasoning capabilities. These findings pave the way for further research and development of more robust and efficient retrieval-augmented generation systems. Our key contributions are as follows:

\input{comparison_table}

\begin{itemize}
    \item We introduce \dataset, a novel dataset of \size test samples designed to evaluate LLMs' ability to retrieve and reason across multiple documents in a unified framework. We plan to open-source this dataset for public use. 
    \item We provide a comprehensive evaluation of state-of-the-art LLMs, highlighting their performance on factuality, retrieval, and reasoning tasks across diverse domains.
    \item We present new empirical insights into the limitations of existing LLMs in handling multi-hop and temporal reasoning tasks, offering avenues for future research to improve these systems.
    \item We propose a multi-step retrieval and reasoning framework that compels models to iteratively retrieve and reason, significantly enhancing their performance on complex queries, from an accuracy of 0.408 with single-step inference to 0.66 with multi-step retrievals.
\end{itemize}

%% file: comparison_table.tex
\begin{table*}[t]
 \small
 \centering
\begin{tabular}
{|p{5.5cm}|c|c|c|c|c|c}
\toprule
\multirow{2}{*}{\centering Dataset} & \multirow{2}{*}{\parbox{1.1cm}{Factuality}} &\multirow{2}{*}{\parbox{1.1cm}{Retrieval}} & \multirow{2}{*}{\parbox{1.3cm}{Reasoning}} &
\multirow{2}{*}{\parbox{1.8cm}{\centering Multi-Hop or Multi-Step}}  & \multirow{2}{*}{\parbox{1.9cm}{\centering Temporal Disambiguation}}  \\
&&&&&\\\midrule
\textbf{\dataset} (our work)& \cmark&\cmark &   \cmark & \cmark & \cmark   \\
\midrule
TruthfulQA~\citep{lin2021truthfulqa} &\cmark &\xmark & \xmark & \xmark & \xmark  \ \\
OpenbookQA~\citep{OpenBookQA2018} &\cmark &\xmark & \xmark & \xmark & \xmark  \ \\
HotpotQA~\citep{yang2018hotpotqa} &\cmark &\xmark & \xmark & \cmark & \xmark  \ \\
HybridQA~\citep{chen2020hybridqa} &  \xmark&\xmark & \cmark & \cmark & \cmark  \\
GSM8k~\citep{cobbe2021gsm8k} &\xmark &\xmark & \cmark & \cmark & \xmark  \\
Multihop-RAG\citep{tang2024multihop} &\cmark &\cmark & \xmark & \cmark & \xmark  \\
MoreHopQA \citep{schnitzler2024morehopqa} &\cmark &\xmark & \cmark & \cmark & \xmark  \\
MuSiQue \citep{DBLP:journals/tacl/TrivediBKS22} &\cmark &\xmark & \cmark & \cmark & \xmark  \\ 
NaturalQuestions~\citep{kwiatkowski2019natural}&\xmark &\cmark & \xmark & \xmark & \cmark  \\
TriviaQA~\citep{joshi2017triviaqa} &\cmark &\xmark & \xmark & \xmark & \xmark  \\
ELI5~\citep{fan2019eli5}&\xmark &\cmark & \cmark & \xmark & \xmark  \\
\bottomrule 
\end{tabular}
\caption{Comparison of \dataset against other datasets. \dataset provides a combination of evaluation samples to test the factuality, retrieval, and reasoning of RAG systems. The dataset also covers multi-hop/step questions along with temporal disambiguation.} 
 \label{tab:dataset_comparison}
\end{table*}

%% file: 030dataset.tex
\section{\dataset}
\vspace{-1mm}

\dataset (\datasetF)  is an evaluation set of 824 questions designed to provide an end-to-end evaluation of Retrieval Augmented Generation (RAG) systems. It assesses three key components of a RAG system: Factuality, Retrieval, and Reasoning. Unlike most existing datasets and benchmarks that evaluate each of these RAG components in isolation, \dataset offers a comprehensive test bed to gain a clear understanding of the overall quality of RAG systems~\citep{DBLP:conf/acl/LinHE22,DBLP:conf/emnlp/Yang0ZBCSM18,DBLP:journals/corr/abs-1710-06481}. This holistic approach allows for a more accurate reflection of how these systems perform in real-world scenarios. In this section, we first detail our data collection process, which involved both synthetic data generation attempts and human annotation. Next, we present the dataset statistics, showcasing the diversity of topics and reasoning types covered. Finally, we outline the rigorous quality checks implemented to ensure the dataset's reliability and challenging nature. By providing this end-to-end evaluation framework, \dataset aims to bridge the gap in existing benchmarks and foster the development of more robust and efficient RAG systems.

\paragraph{Synthetic Data Generation Attempts. } We start our data collection process with synthetic dataset generation to explore a potentially cost-effective alternative to expensive human annotation. We prompt a state-of-the-art LLM with instructions to use multiple articles to generate questions that would require information from multiple articles to answer. The prompt (shown in Figure \ref{tbox:syn_prompt} in the Appendix) takes as input the number of articles provided to generate questions. However, we observed significant issues with this approach. While the \llms were able to generate coherent questions, there was a high proportion of hallucinated questions and answers (>30\%). Additionally, the \llm struggled to generate questions that strictly required more than four articles. To evaluate the potential of this approach, we manually cleaned the hallucinated questions and answers from the obtained set. We then evaluated the same \llm on these cleaned questions and obtained an accuracy of $\sim$32\%, suggesting that the legitimate questions generated by \llms were indeed challenging for state-of-the-art models. There are two key takeaways from our experimentation with synthetic data generation: (1) Synthetic test data requires heavy manual cleaning before usage, which suggests that we will need to rely on human annotations instead of LLMs to generate the final evaluation set; and (2) models performed significantly poorly on the correct test samples we tested on, suggesting that the instruction to create questions can be used to generate a challenging evaluation set.

\input{030reasoning_table}

\paragraph{Human annotation.} Given these findings, we decided to use the core instruction for generating questions that combine information from multiple articles as a guide for human annotation, shown in Figure \ref{tbox:human_annotation_prompt}. This approach aimed to leverage the challenging nature of the synthetic questions while also mitigating the issues of hallucination present in LLM-generated content. Human annotators were tasked with creating questions that required information from multiple Wikipedia articles, following a similar structure to the synthetic prompts but with greater reliability and accuracy. The outcome of this human annotation resulted in \size questions with their correct responses along with the list of Wikipedia articles needed to answer the questions. We also ask the human annotators to label each question based on five reasoning types, i.e, Numerical Reasoning, Tabular Reasoning, Multiple Constraints, Temporal Reasoning, and Post-Processing, described in more details in Table \ref{tab:reasoning}. Please note that a question can belong to multiple reasoning types. To ensure the highest quality annotations, we engaged a team of carefully vetted experts with extensive experience in question generation and complex reasoning tasks.

\begin{figure*}[t!]
    \begin{minipage}[b]{0.5\linewidth}
        \centering
        \includegraphics[width=\textwidth]{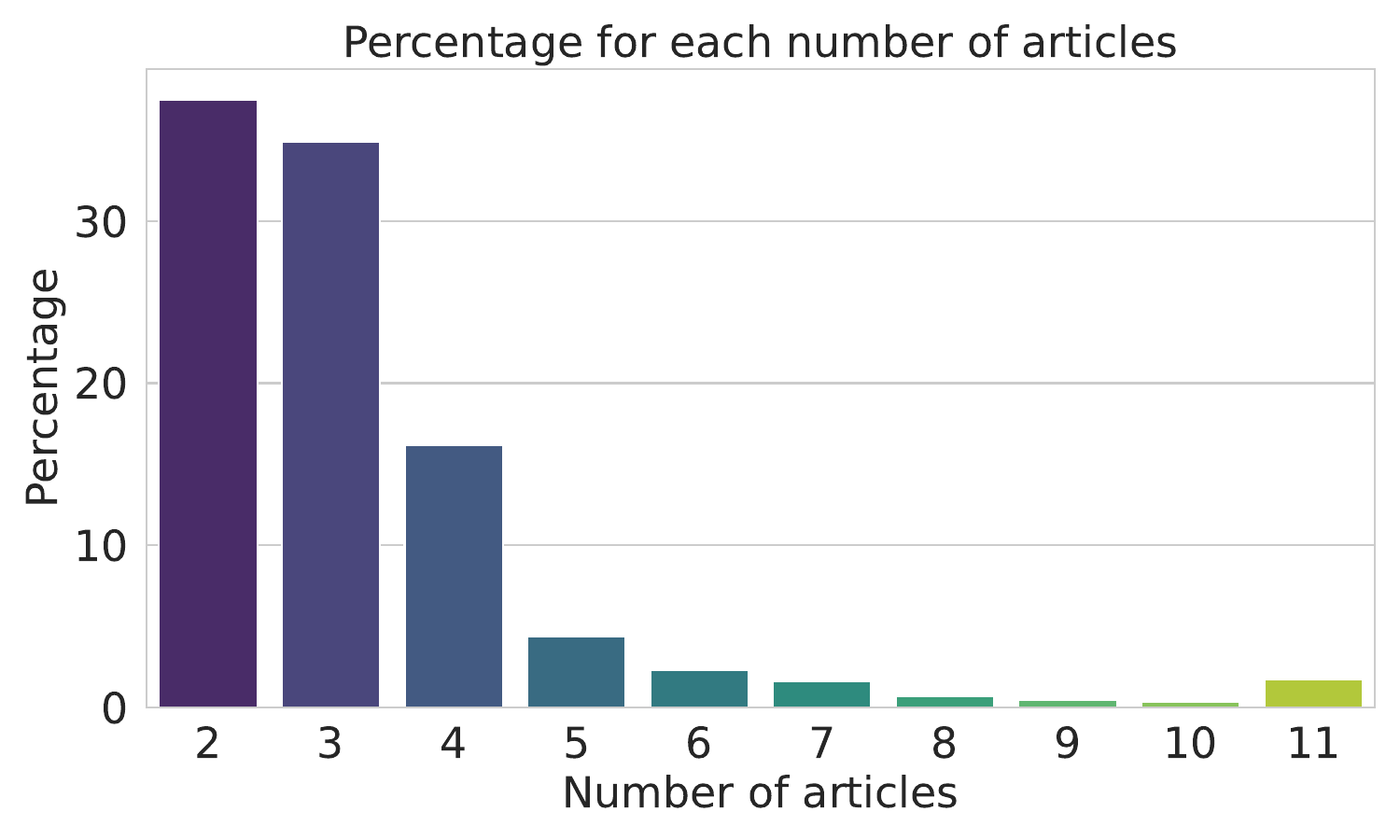}
        \label{fig:figure1}
    \end{minipage}
    \begin{minipage}[b]{0.5\linewidth}
        \centering
        \includegraphics[width=\textwidth]{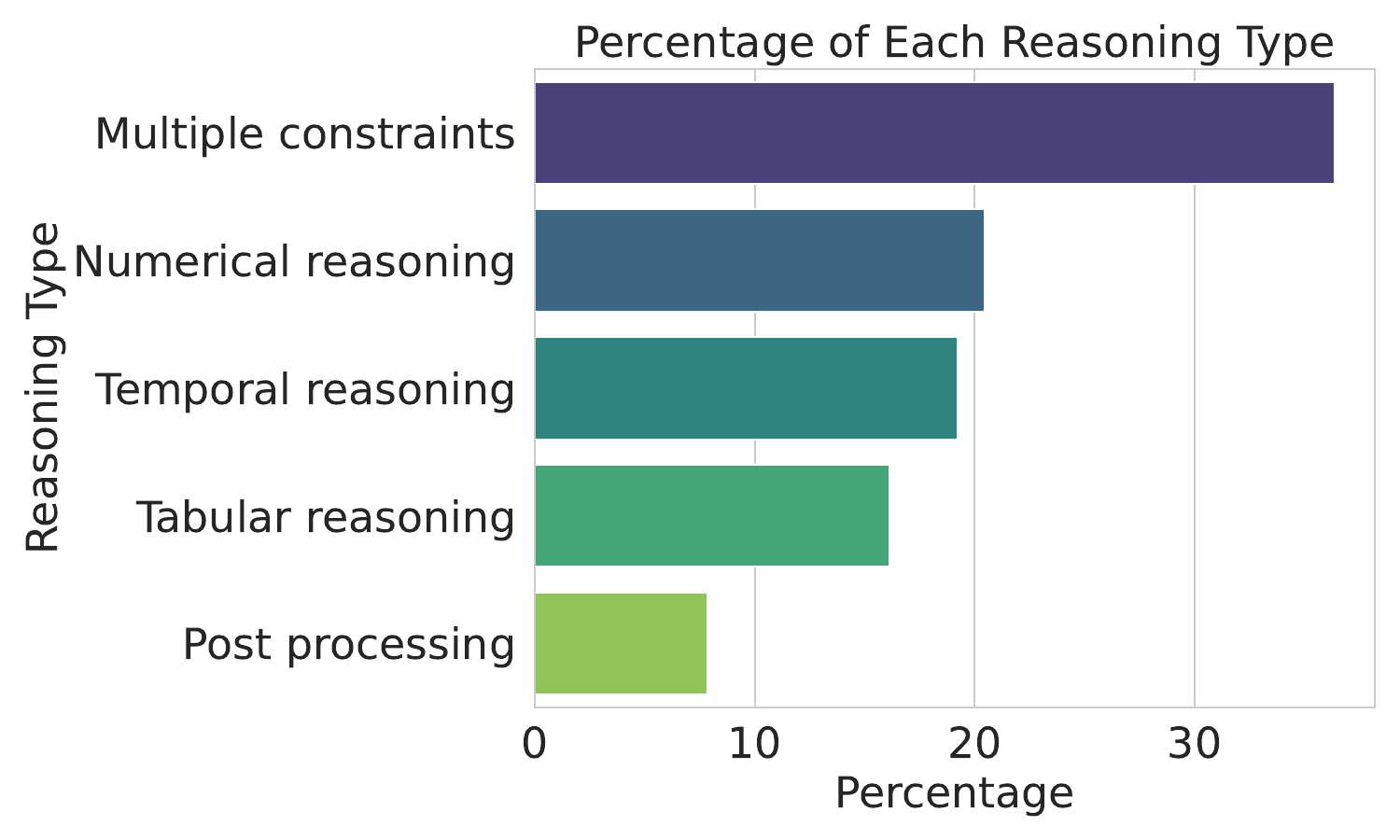}
        \label{fig:figure2}
    \end{minipage}
    \caption{The figure shows the distribution of questions in the \dataset, with the percentage of questions requiring different numbers of Wikipedia articles (left) and the percentage of the dataset belonging to each reasoning type (right). Please note that the percentage bar for 11 denotes the percentage of questions requiring 11 or more Wikipedia articles. }
    \label{data:dist}
\end{figure*}

\paragraph{Dataset Statistics. } The dataset comprises questions related to a diverse set of topics from Wikipedia, involving subjects such as history, sports, science, animals, health, etc. Each question in our dataset require 2-15 Wikipedia articles to answer, with the distribution of the percentage of dataset requiring different numbers of Wikipedia articles shown in Figure \ref{data:dist} (left). Approximately 36\% of questions require two articles to answer, $\sim$35\% require three articles, $\sim$16\% require four articles, and so on. This distribution also represents the general trend of queries asked from LLMs in the real world \citep{liu2009learning}, since the proportion of questions requiring two articles is higher than more complicated questions requiring a greater number of articles. Additionally, we have a healthy distribution of questions belonging to different reasoning types, shown in Figure \ref{data:dist} (right). Questions requiring reasoning over multiple constraints hold the highest percentage of data samples in the test ($\sim$36\%), followed by questions requiring numerical reasoning ($\sim$20\%). Please note that many questions in the dataset also require a combination of different reasoning abilities to find an answer. 

\paragraph{Quality Checks. } Other than the data collection process described in the section above, human annotators also implemented several quality checks to ensure the dataset's high quality and effectiveness in evaluating RAG capabilities:

\begin{itemize}

\item \textbf{Ensuring correctness and groundedness to Wikipedia:}
We verified the correctness of questions and their corresponding answers by re-annotating the questions. Human annotators were asked to confirm if the provided answer was correct and could be answered using the associated Wikipedia pages. This annotation process was conducted three months after collecting the initial data, filtering out 5.5\% of samples where the answer was no longer true after that period.

\item \textbf{Removing ambiguity due to freshness (temporal disambiguation):}
Annotators added extra context to disambiguate answers that could change over time. For example, a question like \textit{"Which country were holders of the FIFA World Cup the last time the UEFA Champions League was won by a club from London?"} was revised to \textit{"As of August 1, 2024, which country were holders of the FIFA World Cup the last time the UEFA Champions League was won by a club from London?"}. This approach mitigates issues with frequent manual updates required for maintaining previous datasets \citep{vu2023freshllms,kasai2024realtime}.

\item \textbf{Preventing guesswork by ensuring a large output space:}
We removed questions with binary answers ("yes" or "no") to prevent LLMs from achieving 50\% accuracy through random guessing. This ensures the dataset is challenging enough to clearly evaluate LLM capabilities.

\item \textbf{Ensuring dataset interpretability and reliability:}
We limited the articles to those from Wikipedia, which has a lower chance of containing unreliable information compared to other sources.

\item \textbf{Addressing data contamination issues:}
To mitigate concerns about potential contamination due to Wikipedia articles being in LLM training sets, we designed questions that require additional reasoning and operations beyond simple fact retrieval. For instance, the question \textit{"How many years earlier would Punxsutawney Phil have to be canonically alive to have made a Groundhog Day prediction in the same state as the US capitol?"} requires not only fact extraction but also additional calculations.

\end{itemize}

%% file: 030reasoning_table.tex
\begin{table*}[t!]
\centering
\small
\begin{tabular}{p{3cm}p{10cm}}
\toprule
\multirow{1}{*}{\textbf{Reasoning Type}} &  \multicolumn{1}{l}{\centering \textbf{Description}} \\
\midrule
Numerical Reasoning & This involves counting, comparisons, or calculations. For example, the question \textit{"How many times faster is the second fastest bird in the Americas compared to the fastest human in the world? Round to the nearest integer."} asks for a calculation comparing the speeds of two objects. \\ 
\midrule
Tabular Reasoning  & This involves statistics found in tables or infoboxes in Wikipedia. For example, the question \textit{"How many runs did the West Indies vice-captain score in the 1983 World Cup?"} requires the answerer to analyze tabular data of top run scorers and extract the relevant information.  \\ 
\midrule
Multiple Constraints & This involves questions with multiple constraints whose intersection points towards a unique answer. For example, \textit{"I'm thinking of an airport near the intersection of US-52 and I-95. Can you remind me which one it is?"} This query has two constraints: first, to locate an airport, and second, that it should be near the intersection of US-52 and I-95. \\ 
\midrule
Temporal Reasoning & This involves reasoning through timelines. For example, \textit{"Leonardo DiCaprio once won an Oscar for best actor. Who won the award for best costume design sixteen years earlier?"}.  \\
\midrule
Post-Processing & This requires the answerer to perform specific post-processing steps after all necessary facts have been gathered. For example, consider the question: \textit{"What is five years after the founding of the largest country in North America in Roman numerals?"}. This question requires the following sub-instructions: (1) Numerical reasoning: Add five years to the founding date, and (2) Post-processing: Convert the resulting year into Roman numerals. \\
\bottomrule
\end{tabular}
\caption{This table provides descriptions of the different reasoning types to which each question in \dataset belongs. The distribution of samples belonging to each reasoning type is shown in Figure \ref{data:dist}.}
\label{tab:reasoning}
\end{table*}

%% file: 040experiments.tex
\vspace{-1mm}
\section{Empirical Analysis}
\vspace{-1mm}

After obtaining a high-quality test set, we evaluate state-of-the-art \llms on their ability to answer questions that require proficiency in factuality, retrieval, and reasoning. Our analysis is divided into two sections: (1) Single-step Evaluations (Section \ref{sec:single-step-evaluation}): Here, we evaluate the \llms based on a single-shot inference, where the idea is to ask the question and assess the response after a single inference call. This evaluation is further divided into cases with and without retrieval to analyze the impact of retrieval on performance. (2) Multi-Step Evaluations (Section \ref{sec:multi-step-evaluation}): In this case, we evaluate the models after making more than a single inference step, focusing on scenarios where retrieval is explicitly required. The motivation for multi-step evaluations is to determine whether forcing the model to retrieve and reason across multiple steps could lead to performance improvements. Next, we describe the details of the two sets of experiments.

\input{results/table/single_step}

\subsection{Single-Step Evaluations}
\label{sec:single-step-evaluation}

In this set of experiments, we evaluate the model using several baseline prompting methods on our test set to understand how well existing \llms perform. Specifically, we experiment with three baseline approaches: \textbf{(1) Naive Prompt:} This is a straightforward approach where we simply ask the question to the model and evaluate if the model’s response without search retrieval contains the correct answer. \textbf{(2) BM25-Retrieved Prompt (n\_docs):} This approach augments the question with the top n\_docs documents having the highest BM25 score \citep{robertson1995okapi} retrieved from a Wikipedia data dump\footnote{\url{https://tinyurl.com/36jxum2y}}. The BM25 score is computed between the question and every article in the Wikipedia dump, after which the top n\_docs with the highest scores are added to the prompt. The motivation behind this approach is to observe improvements in model performance when relevant articles are added to the context. This is denoted as BM25-R (n\_doc) in the results. \textbf{(3) Oracle Prompt:} This prompt includes the question along with all the ground truth Wikipedia articles used by the human annotators to generate the question. The performance of the Oracle Prompt provides the upper bound of model performance in the case of a perfect retrieval system that is able to extract all the relevant articles.

\paragraph{Experimental and Autorater setup. }For the experiments, we use \gemini \citep{geminipro15}, \geminiF \citep{geminiflash15}, \gemmaL \citep{team2024gemma}, \texttt{LLama3.2-3B-I}~\citep{llama}, and \texttt{Qwen2.5-3B-I}~\citep{qwen} as the state-of-the-art \llm since they have shown great performance on several public benchmarks. Since the gold answers to questions in the dataset are free-form tokens instead of choices from multiple-choice answers, we use an \llm to evaluate if the outcome from the \llm under evaluation matches the gold answer, using the prompt shown in Figure \ref{tbox:auto_rate_prompt} in Appendix.  This auto-rating mechanism was tested against human evaluations, in which the \llm-based evaluation showed strong alignment with human annotations (accuracy: 0.96 and Cohen's Kappa: 0.889 for \gemini as autorating \llm), making \llm-based evaluation a suitable approach to evaluate the correctness of model responses.

\begin{figure*}[t!]
    \begin{subfigure}[b]{0.50\linewidth}
        \centering
        \includegraphics[width=\textwidth]{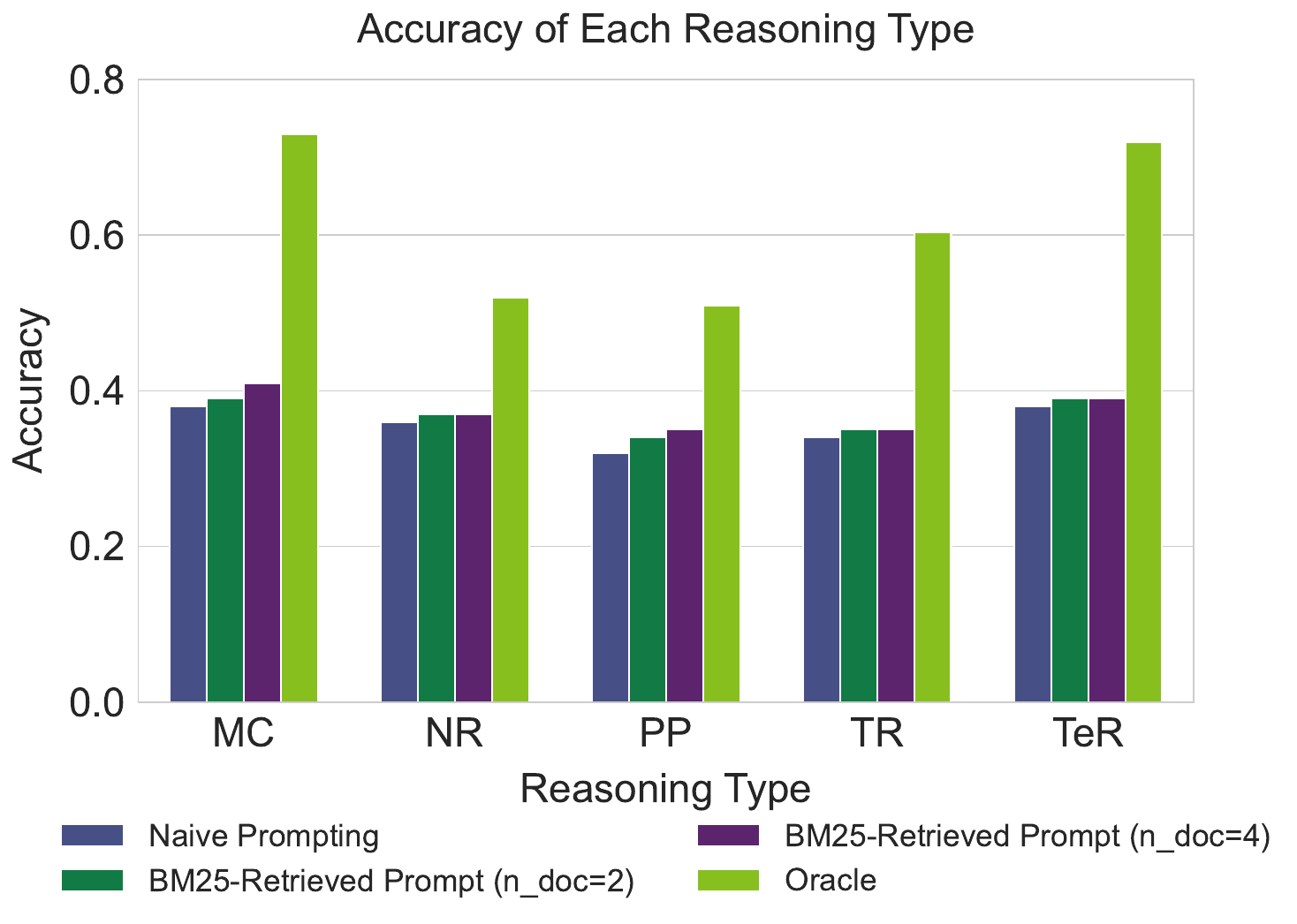}
        \caption{Single-Step Evaluations}
        \label{fig:single_step_e}
    \end{subfigure}
    \hfill
    \begin{subfigure}[b]{0.50\linewidth}
        \centering
        \includegraphics[width=\textwidth]{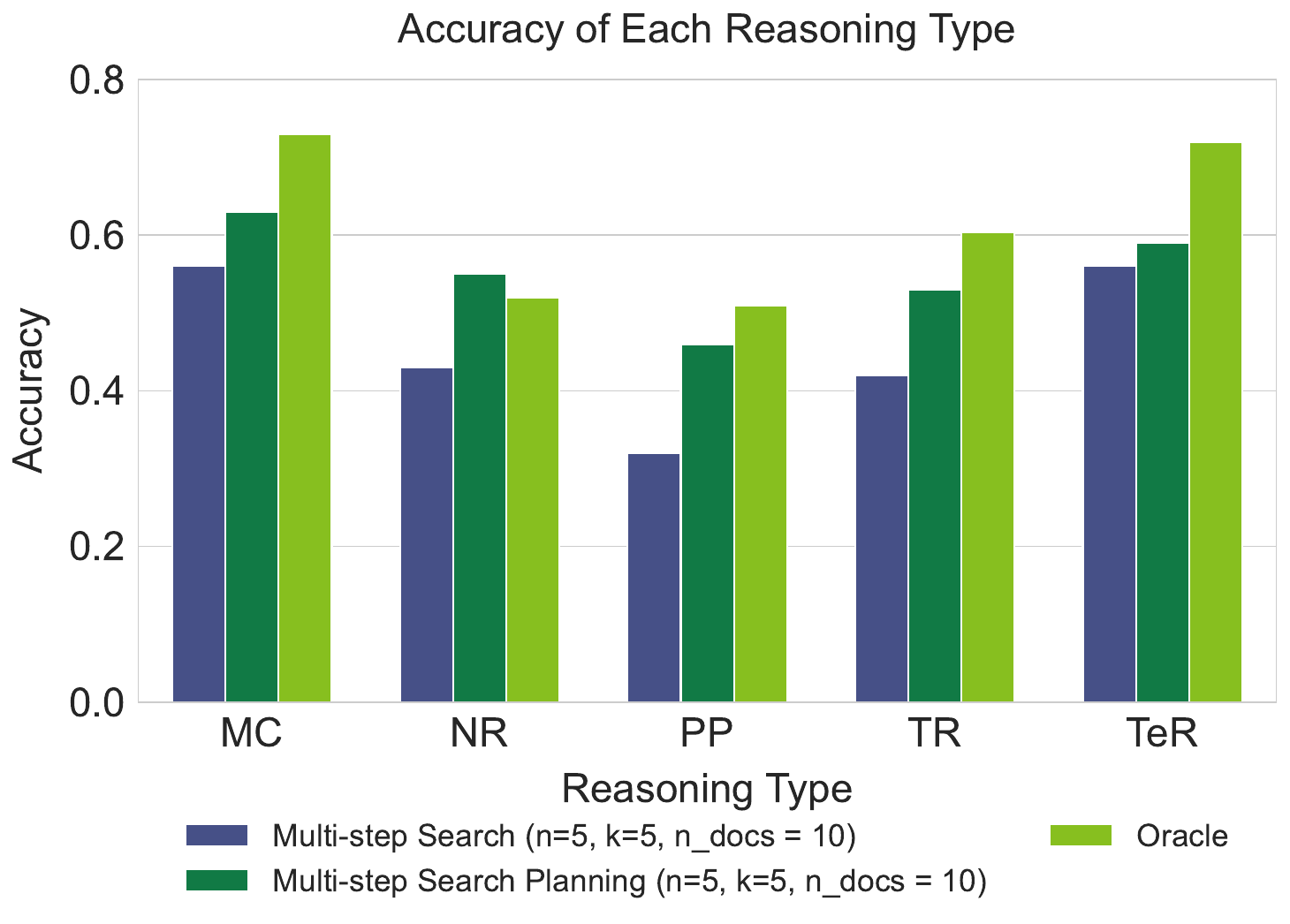}
        \caption{Multi-Step Evaluations}
        \label{fig:multi_step_e}
    \end{subfigure}
    \caption{This plot shows model performance for each reasoning type in single-step (a) and multi-step (b) evaluations. Reasoning types: MC (Multiple Constraints), NR (Numerical Reasoning), PP (Post Processing), TR (Tabular Reasoning), and TeR (Temporal Reasoning). \textbf{(a)} \gemini's accuracy across reasoning types in single-step evaluations. Results show better performance in logical and temporal reasoning, with weaknesses in numerical, tabular, and post-processing reasoning. Oracle information significantly improves accuracy across all categories. \textbf{(b)} \gemini's accuracy in multi-step evaluations. Multi-step search planning increases performance for all reasoning types, with numerical reasoning surpassing oracle performance.}
    \label{fig:reasoning_plot}
\end{figure*}

\paragraph{\llms perform poorly in single-step evaluations. } Based on results shown in Table \ref{tab:baseline}, we observe that naive prompting attains a performance of 0.408 with gradual increases when including BM25 retrieved articles for \gemini. The model achieves an accuracy of 0.452 when the number of documents in the context is 2, and 0.474 when double the number of articles are added to the context. These improvements demonstrate the room for enhancement when the model is able to retrieve relevant articles required to answer the question. The core reason behind these improvements is the improvement in recall in the articles present in context which increased from 0.12 (BM25-R(n\_docs = 2) to 0.15 (BM25-R (n\_docs = 4)). In addition to these approaches, we observe an accuracy of 0.729 for \gemini\ when all the gold Wikipedia articles are provided in the context, which we call Oracle Prompt. Out of 27\% samples where the model made errors, $\sim$80\% of those misclassifications belong to numerical, tabular, and post-processing categories. Hence, these misclassifications show the reasoning gaps in model performance where even after providing all the relevant facts, the model failed to reason through the different facts to provide a correct answer to the question. The accuracies obtained by the Naive Prompt and Oracle Prompt can be considered as the lower bound (when no relevant articles were provided to the model) and upper bound (when all relevant articles were provided to the model) of model performances on \dataset.  This pattern can also be seen in Figure \ref{fig:single_step_e} where we plotted accuracy for each reasoning type and observe that the model performed the lowest in numerical, post-processing, and tabular reasoning tasks. We also observe that adding BM25 retrieved articles primarily helped with questions requiring reasoning through multiple constraints ($\sim$8\% improvement) and post-processing ($\sim$10\% improvement). This aligns well with the fact that providing more relevant articles helps in obtaining facts for each constraint, leading to improvements in performance. We take these learnings and experiment with a more complicated setup where the model is asked to find answer to questions through multiple iterations instead of a single step. 

\begin{algorithm}[t]
\caption{Multi-Step Evaluation with BM25 Retrieval}
\label{alg:multi-step-bm25-unique}
\begin{algorithmic}[1]
\STATE \textbf{Input:} Initial question $Q$, number of iterations $n$, number of queries $k$, number of documents $n\_docs$
\STATE \textbf{Output:} Final response $R$
\STATE Initialize context $C \gets \{Q\}$
\FOR{iteration $i = 1$ to $n$}
    \FOR{query $j = 1$ to $k$}
        \STATE Generate search query $Q_{ij}$ based on context $C$
        \STATE Retrieve top $n\_docs$ documents $D_{ij}$ using BM25 based on $Q_{ij}$
        \STATE $C \gets C \cup (D_{ij} \setminus C)$  \COMMENT{Add only new documents to context}
    \ENDFOR
\ENDFOR
\STATE Generate final response $R$ using context $C$
\STATE \textbf{return} $R$
\end{algorithmic}
\end{algorithm}

\subsection{Multi-Step Evaluations}
\label{sec:multi-step-evaluation}

Based on the findings from the previous experiment with single-step evaluations, where we observed an increase in performance when related articles are added to the context, we were led to explore a setting where the model is compelled to plan its search for relevant Wikipedia articles in order to find answers. More specifically, we design a pipeline where the model is asked a question along with the instruction to generate $k$ search queries which are then used to extract the top-n\_docs Wikipedia articles with the highest BM25 scores. These documents are then appended to the context. This process of query generation and retrieved article augmentation is carried forward for $n$ steps. Once the $n$ steps of retrieval are completed, the model is asked to answer the question based on the articles appended in the context, as shown in Algorithm \ref{alg:multi-step-bm25-unique}. We conduct two sets of experiments here: (1) Vanilla with no explicit planning instructions, and (2) With search planning instructions to help the model navigate the search process efficiently. To implement this pipeline, we used the simplest document retrieval component which is essentially an index of Wikipedia pages, where the articles with the highest BM25 scores for each query are returned to the LLM and added to the context. This retrieval component is used instead of making direct calls to an online search engine for two reasons: (1) We would like to keep the retrieval system constant to clearly evaluate the search planning capability of the LLMs instead of the retrieval system's capability in returning the most relevant articles, and (2) The BM25-based retrieval system makes our pipeline reproducible and limits the search space to Wikipedia pages only, as the questions were generated from Wikipedia articles only.

\begin{figure*}[t!]
    \begin{minipage}[b]{0.48\linewidth}
        \centering
        \includegraphics[width=\textwidth]{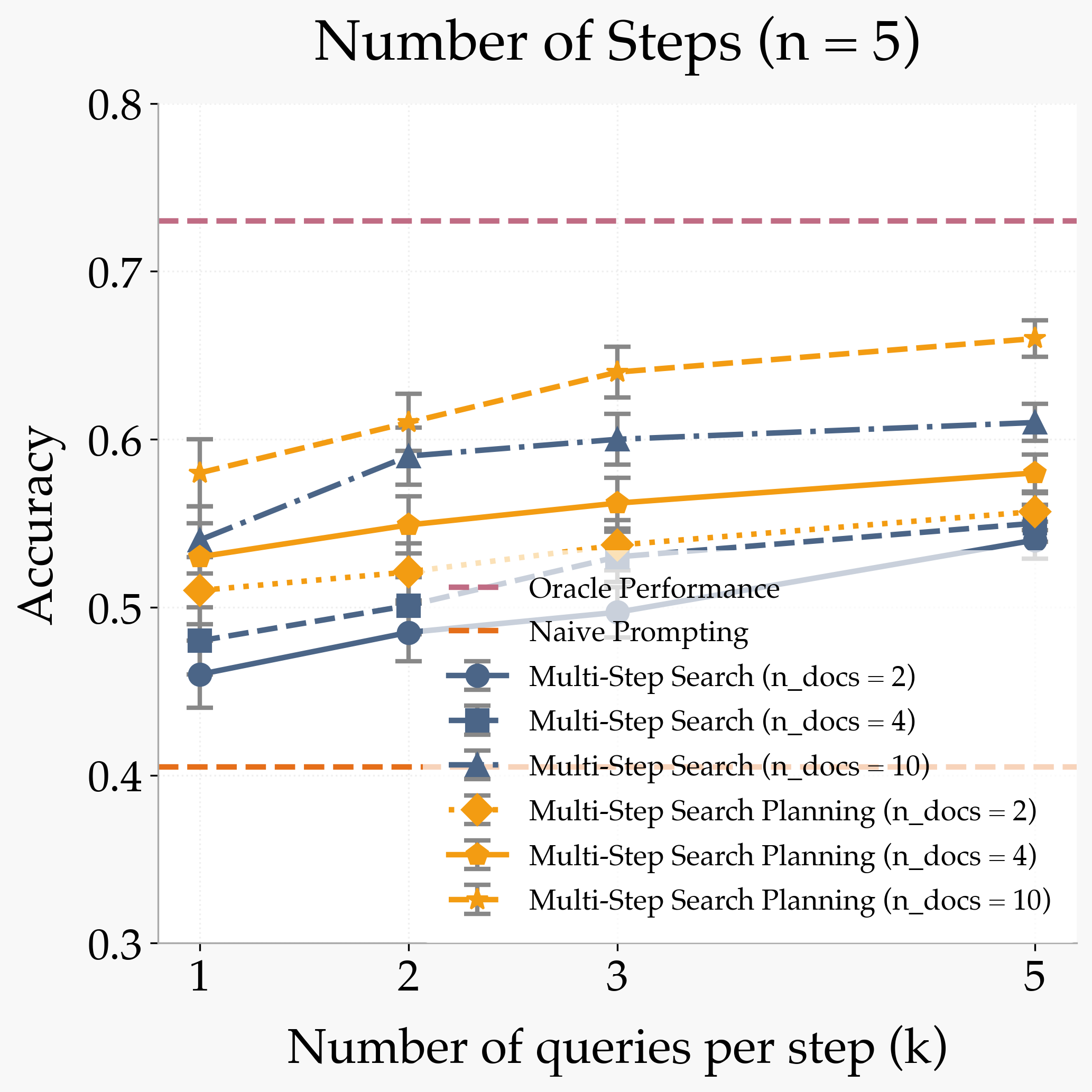}
        \label{fig:figure1}
    \end{minipage}
    \begin{minipage}[b]{0.48\linewidth}
        \centering
        \includegraphics[width=\textwidth]{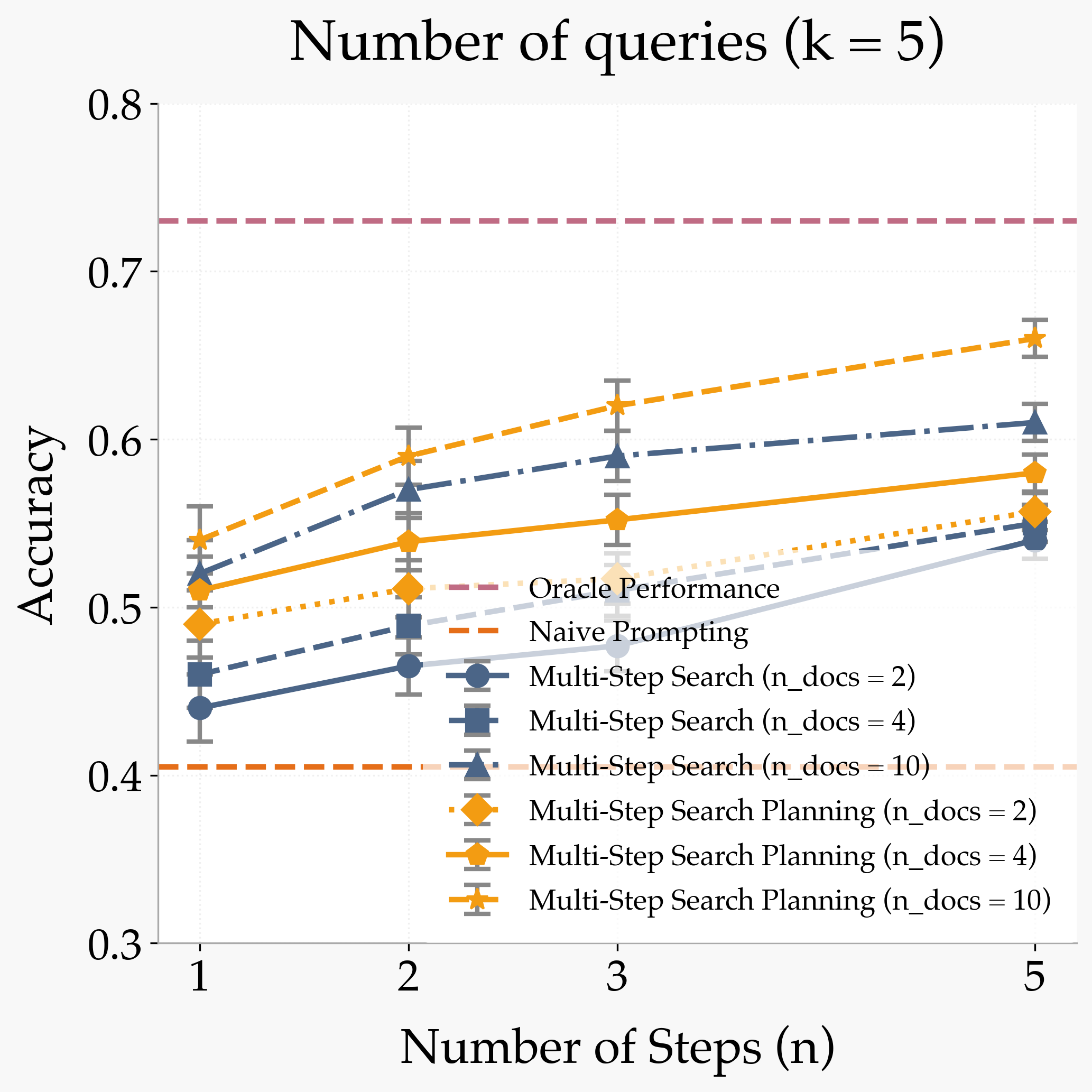}
        \label{fig:figure2}
    \end{minipage}
    \caption{The figure shows the performance improvements when the number of steps (n) and number of queries per step (k) is changed. We achieved the best performance of 0.66 with the combination of ($k$, $n$, $n\_docs$) = (5,5,10)} 
    \label{fig:multi_step_1}
\end{figure*}

\paragraph{Multi-Step retrieval significantly improves model performance. } We plot model performance on different combinations of ($k$, $n$, $n\_docs$) in Figure  \ref{fig:multi_step_e}. Based on these results, we observe a steady increase in performance as the number of steps and queries are increased with accuracy increasing from $\sim0.45$ to $\sim0.52$ for the case of ($k$, $n$, $n\_docs$)=(5,5,2) for vanilla setting where the model is not provided with any specific planning instructions. This is expected, as more steps and queries allow the model to add more relevant documents to its context, leading the model to improve recall, which translates to better accuracy. However, the performance still remains quite low even after five iterations of search retrievals, which is computationally expensive since this requires six non-parallelizable inference calls (five for retrieval + one for final answering) to answer each question. One of the reasons we found behind this slow progress in performance is the lack of diversity in the queries generated by the model; it seemed like the model goes in the wrong direction in search retrievals and never corrects itself. To mitigate this problem, we experiment with two changes to the instructions: (1) We provide a few examples of how an ideal best-case search query sequence should look, and (2) We provide instructions not to repeat queries and force the model to \textit{"think step-by-step"} \citep{kojima2022large}. We observe a very promising trend with these changes, where the model performance (0.66) through iterations reaches close to the oracle performance (0.73) by the end of five iterations of retrievals, as shown in Figure \ref{fig:multi_step_1}. We hope our benchmark will be useful for the community to further reduce the number of search calls and improve model accuracy.

%% file: results/table/single_step.tex
\begin{table*}[t!]
\centering
\small
\begin{tabular}{p{3cm}*{5}{r}}
\toprule
\multirow{1}{*}{Baselines} &  \multicolumn{1}{c}{\texttt{G-Pro-1.5}} & \multicolumn{1}{c}{\texttt{G-Flash-1.5}}   & \multicolumn{1}{c}{\gemmaL} & \multicolumn{1}{c}{\texttt{LLama3.2-3B-I}} & \multicolumn{1}{c}{\texttt{Qwen2.5-3B-I}} \\
Date of Release & 5/14/24 & 5/14/24 & 6/27/24 & 9/25/24 & 9/19/24 \\
\midrule
Naive Prompt & 0.408 & 0.263 & 0.308 &  0.115 & 0.095 \\  
BM25-R (n$\_$doc=2)  & 0.452 & 0.288  & -- & -- & --   \\ 
BM25-R (n$\_$doc=4) & 0.474 & 0.315 &  -- & -- & --\\ 
Oracle Prompt & 0.729 & 0.665 & -- & -- & -- \\ 
\bottomrule
\end{tabular}
\caption{This table presents the accuracy performance of \gemini (G-Pro-1.5), \geminiF (G-Flash-1.5), \gemmaL, \texttt{LLama3.2-3B-I} and \texttt{Qwen2.5-3B-I} on our proposed evaluation dataset. Please note that the performance of Gemma, Llama, and Qwen models is not reported for cases requiring longer context due to the small maximum context length of the model.}
\label{tab:baseline}
\end{table*}

%% file: 020relwork.tex
\section{Related Works}
\vspace{-2mm}
Evaluating Retrieval-Augmented Generation (RAG) systems is crucial as they combine retrieval and generative capabilities~\citep{yu2024evaluation}. Existing benchmarks like NaturalQuestions~\citep{kwiatkowski2019natural}, TriviaQA~\citep{joshi2017triviaqa}, and ELI5~\citep{fan2019eli5} focus on specific aspects but lack comprehensive assessment. NaturalQuestions tests retrieval precision, TriviaQA emphasizes factual correctness, and ELI5 targets explainability without rigorous multi-hop reasoning evaluation.

\dataset addresses these limitations by offering a unified evaluation framework. It tests RAG systems across factual retrieval, reasoning, and synthesis dimensions. Unlike existing datasets, \dataset incorporates complex multi-hop queries requiring information retrieval and integration from various sources, while handling temporal disambiguation. It also assesses information synthesis into coherent responses, evaluating RAG systems in realistic, multifaceted scenarios. This makes \dataset a more rigorous and comprehensive benchmark for guiding next-generation RAG system development. Additional dataset comparisons are provided in Table \ref{tab:dataset_comparison}.

%% file: 050conclusions.tex
\section{Conclusion}

In this work, we introduced \dataset, a comprehensive evaluation dataset designed to test the capabilities of Retrieval-Augmented Generation (RAG) systems across factuality, retrieval accuracy, and reasoning. Our experiments with state-of-the-art \llms highlight the existing gaps in their ability to handle complex, multi-hop reasoning tasks. The baseline results showed that even advanced models struggle significantly with the challenging scenarios presented in \dataset, achieving only moderate improvements when multi-step retrieval and reasoning strategies were employed. The \dataset dataset addresses a critical need in the evaluation of RAG systems by offering an integrated framework that tests these systems in a more holistic manner compared to existing benchmarks. By simulating realistic, multi-document queries, \dataset provides a clearer picture of the current capabilities and limitations of \llms in real-world applications. Our findings underscore the importance of further enhancing both the retrieval mechanisms and the reasoning capabilities of these models to improve their overall performance.

\section*{Limitations}
While \dataset provides a comprehensive evaluation framework for RAG systems, it is important to acknowledge certain limitations. One significant concern is the potential for pretraining data contamination. As large language models are trained on vast amounts of internet data, there is a risk that some of the information in our dataset may have been seen by these models during their pretraining phase. This could lead to artificially inflated performance metrics and reduce the dataset's effectiveness in measuring true generalization capabilities. Future iterations of \dataset should explore techniques to mitigate this issue, such as using more recent or synthetic data, or developing methods to quantify and account for potential contamination. Additionally, while we have strived for diversity in our dataset, it may not fully represent the entire spectrum of real-world queries and scenarios, potentially limiting its applicability to certain domains or use cases. Addressing these limitations will be crucial for improving the robustness and reliability of RAG system evaluations.

\section*{Ethical Considerations}
In developing \dataset, we recognize the importance of addressing potential ethical considerations in retrieval-augmented generation (RAG) research. Researchers using this benchmark should be mindful of possible biases in information sources, the need for safeguards against potential misuse of advanced RAG systems, and the environmental impact of computationally intensive approaches. We encourage exploring efficient, accessible implementations and maintaining transparency in methodologies. By openly discussing these considerations, we aim to foster responsible advancement of RAG capabilities. We welcome ongoing dialogue within the research community to collectively navigate these important ethical aspects as the field progresses.

\section*{Acknowledgements}
We thank Abhimanyu Goyal, Tsendsuren Munkhdalai, and Dipanjan Das for helpful feedback on writing and presentation of this work.

%% file: appendix.tex
\clearpage
\input{synthetic_dg_prompt}

\input{auto_rate_prompt}

\input{031userstudy}

\section{Future Work}
 Moving forward, there are several promising avenues for future research. First, the development of more sophisticated retrieval strategies is essential. This includes exploring dense retrievers trained directly on the multihop retrieval task, such as those based on ColBERT \citep{khattab2020colbert}, or SimCSE \citep{gao2021simcse} architectures. These approaches could better handle diverse and complex queries by adapting to the context iteratively. Second, improving the reasoning capabilities of \llms remains a significant challenge. We can explore process supervision methods like those used in PRM-800K \citep{lightman2023let}, or investigate distillation techniques on successful trajectories, similar to approaches in ToolFormer \citep{schick2024toolformer} and DSPy~\citep{khattab2023dspy}. These methods could enhance numerical, temporal, and post-processing reasoning. Additionally, we can explore modeling approaches such as context reduction of wiki articles to improve planning capabilities and training query generators for more effective information retrieval. Lastly, expanding the \dataset dataset to include more diverse and domain-specific questions, as well as incorporating more dynamic elements such as real-time information retrieval, could further enhance its utility as a benchmark for next-generation RAG systems. It is important to note that future work should also address the potential limitations of our current approach, including the risk of pretraining data contamination, which may affect the generalizability and reliability of the results, particularly when using Wikipedia articles that could overlap with LLM training data.


%% file: synthetic_dg_prompt.tex
\begin{figure*}[h]
\begin{tcolorbox}[colframe=myblue, title=\textbf{Synthetic Data Generation}]
\textbf{\texttt{System:}} You are a helpful assistant.\\
\textbf{\texttt{User:}} """\textbf{TASK:} You will be provided with \texttt{\{k\_context\}} Wikipedia article extracts. Based on these extracts, generate \texttt{\{n\_questions\}} challenging factoid questions that meet the following criteria: \\
1. \textbf{Standalone \& Context-Independent:} Questions should not contain any references to "Article 1", "Article 2", etc. They should be understandable without any additional context. \\
2. \textbf{Unambiguous Answer:} Each question should have a single, clear, and factual answer. \\
3. \textbf{Multi-hop Reasoning:} Answering each question should require combining information from ALL \texttt{\{k\_context\}} provided Wikipedia articles. \\
4. \textbf{Grounded in Context \& Conceptual Format:} Each question must conceptually follow this format, seamlessly integrating information from each article: \\
**Start with a clear question word (What/How/Where/When).**  \\
**Introduce information from each article step-by-step, using connectors to link them logically.** \\
**Example connectors: 'in relation to', 'compared to', 'as a result of', 'which also', 'in addition to'.  \\
** For each question: *  \\
**Provide the single-word answer in parentheses after the question mark.** * \\
**On a new line, clearly explain the reasoning process.** * \\
**For each article, bullet point the specific piece of information used to formulate the question.** \\ \\
\textbf{Example:} \\
**Question:** What type of bird, belonging to the Ardeidae family, went extinct around 1690 and was known for its terrestrial abilities? (Dodo) \\
**Reasoning:** * 
**Article 1:** Provides information about the Dodo belonging to the Ardeidae family. * 
**Article 2:** Mentions the extinction of the Dodo around 1690. * 
**Article 3:** Highlights the Dodo's adaptation to terrestrial life. \\ \\ \texttt{\{WIKI ARTICLES\}}
""" 
\end{tcolorbox}
\caption{Prompt used to generate questions synthetically using \gemini. \texttt{{k\_context}} and \texttt{{n\_questions}} are placeholders for the number of articles provided and the number of questions to generate per inference.}
\label{tbox:syn_prompt}
\end{figure*}

%% file: auto_rate_prompt.tex
\begin{figure*}[h]
\begin{tcolorbox}[colframe=myblue, title=\textbf{Auto-rating Prompt}]
\textbf{\texttt{System:}} You are a helpful assistant.\\
\textbf{\texttt{User:}} """===Task===\\ \\
    I need your help in evaluating an answer provided by an LLM against a ground truth answer. Your task is to determine if the ground truth answer is present in the LLM’s response. Please analyze the provided data and make a decision.
\\
    ===Instructions=== \\
    1. Carefully compare the "Predicted Answer" with the "Ground Truth Answer". \\
    2. Consider the substance of the answers – look for equivalent information or correct answers. Do not focus on exact wording unless the exact wording is crucial to the meaning. \\
    3. Your final decision should be based on whether the meaning and the vital facts of the "Ground Truth Answer" are present in the "Predicted Answer:" \\

    ===Input Data=== \\
    - Question: \texttt{<<question>>} \\
    - Predicted Answer: \texttt{<<LLM\_response>>} \\
    - Ground Truth Answer: \texttt{<<ground\_truth\_answer>>}\\

    ===Output Format=== \\
    Provide your final evaluation in the following format: \\
    "Explanation:" (How you made the decision?) \\
    "Decision:" ("TRUE" or "FALSE" ) \\
\\
    Please proceed with the evaluation. 
""" 
\end{tcolorbox}
\caption{Prompt used to auto-rate the responses of \llm in the experiments. The \llm is provided with questions, model responses, and ground truth answers, along with instructions to check if the model response contains the gold answer.}
\vspace{-10pt}
\label{tbox:auto_rate_prompt}
\end{figure*}

%% file: 031userstudy.tex
\begin{figure*}[h]
\begin{tcolorbox}[colframe=myblue, title=\textbf{Task Instruction Prompt for Human Annotation}]
\textbf{Task Description}

Create \texttt{n} factoid questions that demand multi-hop reasoning based on information found across multiple Wikipedia articles. These questions should ideally have a single, unambiguous answer and may optionally incorporate elements of challenging reasoning.
 
Here's a breakdown of the key terms and their relationships:
\begin{itemize}
    \item \textbf{Factoid Questions:} These are trivia-style questions with a single, clearly defined, and factually correct answer.
    \item \textbf{Multi-hop Reasoning:} This refers to the core requirement that answering the questions necessitates combining information from different sections within multiple Wikipedia articles. The example provided ("What is the name of the river that flows through the city where the Eiffel Tower is located?") highlights how this differs from a simple factoid question ("What is the capital of France?").
    \item \textbf{Challenging Reasoning:} This aspect encourages the inclusion of questions that go beyond simple information retrieval and demand critical thinking. This can be achieved through various question types like:
    \begin{itemize}
        \item Numerical Reasoning:  Involving counting, comparisons, or calculations. \texttt{<examples>}
        \item Tabular Reasoning: Involving statistics found in tables / info boxes in wikipedia. \texttt{<examples>}
        \item Multiple Constraints: Questions involving multiple constraints, whose intersection points towards a unique answer. \texttt{<examples>}
        \item Temporal Reasoning : Questions involving reasoning through timelines. \texttt{<examples>}
        \item Post processing: This involves requiring the answerer to perform specific post-processing steps after all necessary facts have been gathered. \texttt{<examples>}
    \end{itemize}
\end{itemize}

Additional Requirements:
\begin{itemize}
    \item Wikipedia Articles:  The information used to answer the questions must be sourced from Wikipedia articles.
    \item Standalone and Context-Independent: Questions should be understandable without requiring additional information or context.
    \item Single, Unambiguous Answer: Each question should have only one correct answer, leaving no room for ambiguity.
    \item Avoid boolean questions (yes/no questions) that can be answered with a simple "yes" or "no." \texttt{<examples>}
\end{itemize}
\end{tcolorbox}
\caption{Task instruction provided to human annotators to generate samples for \dataset.}
\vspace{-10pt}
\label{tbox:human_annotation_prompt}
\end{figure*}